\documentclass{article}

\usepackage[preprint]{neurips_2026}
\setcitestyle{numbers,square,comma}  

\usepackage{amsmath,amssymb}
\usepackage{booktabs}
\usepackage{xcolor}
\usepackage{graphicx}
\usepackage{microtype}
\usepackage{enumitem}
\usepackage{pifont}
\usepackage{multirow}
\usepackage{subcaption}
\usepackage[colorlinks=true,linkcolor=blue!60!black,citecolor=green!50!black,
            urlcolor=blue!60!black]{hyperref}
\graphicspath{{figures/}}

\makeatletter
\renewcommand{\paragraph}{\@startsection{paragraph}{4}{\z@}%
  {0.4ex \@plus 0.1ex \@minus 0.1ex}{-1em}{\normalsize\bf}}
\renewcommand{\section}{\@startsection{section}{1}{\z@}%
  {0.8ex \@plus 0.2ex \@minus 0.2ex}{0.4ex \@plus 0.1ex}{\large\bf}}
\renewcommand{\subsection}{\@startsection{subsection}{2}{\z@}%
  {1.0ex \@plus 0.2ex \@minus 0.1ex}{0.3ex \@plus 0.1ex}{\normalsize\bf}}
\makeatother
\title{Same Brain, Different Prediction:\\How Preprocessing Choices Undermine EEG Decoding Reliability}

\author{
  Dengzhe Hou\textsuperscript{1,$\dagger$} \quad
  Zihao Wu\textsuperscript{2} \quad
  Lingyu Jiang\textsuperscript{1} \quad
  Zirui Li\textsuperscript{1} \quad
  Fangzhou Lin\textsuperscript{1,3,4} \quad
  Kazunori D Yamada\textsuperscript{1} \\
  \textsuperscript{1}Tohoku University \quad
  \textsuperscript{2}University of Georgia \\
  \textsuperscript{3}Texas A\&M University \quad
  \textsuperscript{4}Worcester Polytechnic Institute \\
  \texttt{dengzhe.hou.a5@tohoku.ac.jp} \\
  \textsuperscript{$\dagger$}Corresponding author.
}

\begin{document}
\maketitle

\begin{abstract}
Electroencephalography (EEG) is a cornerstone of brain-computer interfaces and clinical neuroscience, yet deep learning models are typically trained and evaluated under a single, unreported preprocessing pipeline. We formalize preprocessing choices as a counterfactual intervention space and show that EEG predictions are surprisingly unstable under this space: across six datasets spanning four paradigms, \textbf{up to 42\% of trial-level predictions flip} when only the preprocessing changes, a variability that standard uncertainty methods do not explicitly quantify because they condition on a fixed preprocessing pipeline.
We provide three tools to make this instability measurable, decomposable, and reducible. First, a Walsh-Hadamard decomposition of the $2^7$ pipeline space reveals that sensitivity is near-additive in practice under the binary intervention design, enabling efficient step-by-step optimization. Second, we introduce \textbf{Preprocessing Uncertainty} (PU), a per-trial diagnostic that captures a dimension of instability complementary to model-based confidence. Third, we study \textbf{Normalized Adaptive PGI} (NA-PGI), a graph-structured regularizer that exploits the compositional structure of preprocessing interventions as one mitigation strategy with clear scope conditions. Code is available at \url{https://github.com/dengzhe-hou/EEG-Preprocessing-Sensitivity}.
\end{abstract}

\section{Introduction}
\label{sec:intro}

Electroencephalography (EEG) is among the lowest signal-to-noise ratio (SNR) modalities in neuroscience: single-trial cortical responses are on the order of microvolts, routinely buried under physiological and environmental noise orders of magnitude larger. Despite this, a growing body of EEG deep learning research~\citep{roy2019deep,craik2019deep} reports confident predictions, often without acknowledging a hidden source of variability that existing uncertainty methods cannot capture: the preprocessing choices made by the analyst. Recent EEG foundation models~\citep{kostas2021bendr,jiang2024large,yang2023biot,csbrain2025,neuript2025,reve2025} learn cross-dataset representations from thousands of subjects, but their robustness to preprocessing variation remains unexamined.

Every EEG study involves at least seven independent preprocessing decisions, reference scheme, high-pass cutoff, low-pass cutoff, baseline correction, artifact attenuation, epoch rejection, and bad-channel repair, each with commonly used alternatives. These choices are rarely justified, rarely reported in full, and never tested for their impact on individual predictions. We show that this oversight has severe consequences: across six datasets spanning motor imagery, sleep staging, event-related potentials, and emotion recognition, \textbf{up to 42\% of trial-level predictions flip} when only the preprocessing changes, the model, the data, and the labels remain identical.

This finding exposes a reliability gap in EEG deep learning. Standard uncertainty methods (softmax entropy, MC Dropout~\citep{gal2016dropout}, deep ensembles~\citep{lakshminarayanan2017simple}) hold preprocessing fixed and therefore do not quantify this source of instability. A decoder may report 95\% confidence on a trial that would receive the opposite prediction under an equally valid pipeline.

We formalize preprocessing choices as a \textbf{counterfactual intervention space} and provide three tools to make the resulting instability measurable, decomposable, and reducible:

\begin{enumerate}[nosep,leftmargin=*]
\item \textbf{Decomposition.} A Walsh-Hadamard analysis of the $2^7$ accuracy hypercube reveals that sensitivity is near-additive in practice under the binary intervention design ($\leq$0.2\% of total variance in interactions); greedy step-by-step optimization achieves accuracy within 2.5\% of the oracle on all six datasets.
\item \textbf{Diagnostics.} Preprocessing Uncertainty (PU), a per-trial measure of pipeline disagreement, correlates only moderately with softmax entropy ($\rho{=}0.40$) and MC Dropout ($\rho{=}0.33$), capturing an otherwise invisible dimension of instability. Signal-level analysis links step-level sensitivity to measurable properties (e.g., kurtosis reduction from high-pass filtering, $r{=}0.58$).
\item \textbf{Mitigation.} Normalized Adaptive PGI (NA-PGI), a graph-structured regularizer exploiting the compositional lattice of preprocessing interventions, reduces CFR by up to 35\% with a single transferable hyperparameter ($\lambda{=}1$), studied here as one mitigation strategy with clear scope conditions.
\end{enumerate}

\begin{figure}[t]
    \centering
    \includegraphics[width=\linewidth]{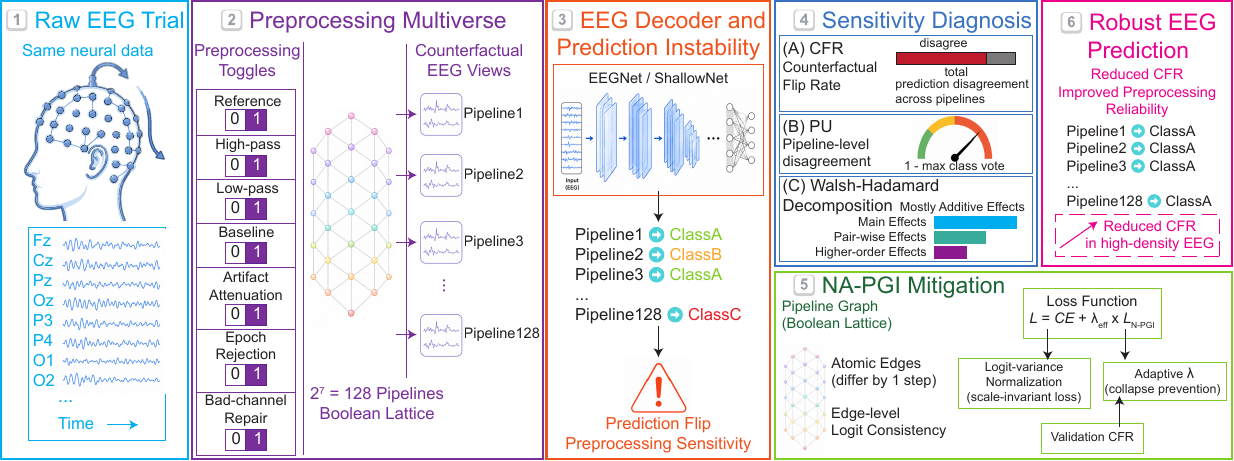}
    \caption{\textbf{Overview.} (1)~A raw EEG trial is processed through (2)~seven binary preprocessing toggles forming a $2^7{=}128$-pipeline Boolean lattice, producing 128 counterfactual views. (3)~An EEG decoder (EEGNet/ShallowNet) yields different predictions across pipelines, exposing preprocessing sensitivity. (4)~Three diagnostics quantify this instability: CFR (flip rate), PU (per-trial pipeline disagreement), and Walsh-Hadamard decomposition (additive structure). (5)~NA-PGI mitigates sensitivity via edge-level logit consistency with logit-variance normalization and adaptive $\lambda$. (6)~The result is more preprocessing-stable prediction, particularly in high-density EEG settings.}
    \label{fig:overview}
\end{figure}


\section{Related Work}
\label{sec:related}

\paragraph{EEG preprocessing pipelines.}
Standardized EEG preprocessing pipelines include PREP~\citep{bigdely2015prep}, DISCOVER-EEG~\citep{pedroni2023discover}, RELAX~\citep{bailey2023relax}, and CLEAN~\citep{clean2026}. These pipelines codify expert knowledge into fixed recipes but do not optimize for downstream decoding performance. All are MATLAB-based and rule-driven, with no mechanism to adapt to the task or model.

\paragraph{Preprocessing impact on deep learning.}
\citet{delpup2025more} trained 4,800 models across 6 tasks and found that minimal preprocessing (filtering only, no artifact removal) often outperforms complex pipelines for deep learning, suggesting that artifacts may carry useful information. \citet{kessler2025preprocessing} systematically varied filtering, referencing, and artifact correction and showed that these choices can reverse model rankings. Neither study decomposed the contribution of individual steps or compared across tasks. Crucially, all prior preprocessing studies compare \textbf{pipeline-level performance} (which pipeline yields higher accuracy); our unit of analysis is the \textbf{counterfactual prediction of the same raw trial} across preprocessing choices, shifting the question from ``which preprocessing is better?'' to ``how reliable is any single prediction?'' More broadly, shortcut learning~\citep{geirhos2020shortcut} and underspecification~\citep{damour2022underspecification} suggest that preprocessing choices may create systematic biases that models exploit without generalization, a concern amplified by recent evidence that deep networks can memorize arbitrary label assignments~\citep{zhang2021understanding}.

\paragraph{Multiverse analysis and pipeline-invariant learning.}
The ``garden of forking paths'' framework~\citep{steegen2016increasing} recognizes that analytical flexibility inflates false positives.
\citet{botvinik2020variability} showed that 70 fMRI teams reached divergent conclusions from the same data; \citet{eegmultiverse2025} computed 528 EEG pipelines but focused on statistical analysis.
\citet{li2022pipeline} proposed pipeline-invariant learning for MRI, treating pipelines as opaque domains. Recent fMRI foundation models such as NeuroSTORM~\citep{wang2026neurostorm} aim to learn preprocessing-robust representations through large-scale pretraining on 50,000+ participants, but do not quantify residual preprocessing sensitivity. This connects to the broader underspecification problem in ML~\citep{damour2022underspecification}, where multiple pipelines achieve similar validation performance but diverge at deployment. We differ from all prior work by decomposing pipelines into atomic interventions and applying factorial analysis to EEG decoding, providing a principled diagnostic for any model's preprocessing robustness.

\paragraph{EEG foundation models.}
A wave of EEG foundation models has emerged: BENDR~\citep{kostas2021bendr} applies contrastive pretraining, LaBraM~\citep{jiang2024large} scales to large multi-dataset corpora, BIOT~\citep{yang2023biot} addresses cross-modal biosignal learning, CSBrain~\citep{csbrain2025} introduces cross-scale spatiotemporal modeling across 16 datasets, NeurIPT~\citep{neuript2025} uses mixture-of-experts for heterogeneous EEG, and REVE~\citep{reve2025} pretrains on 25,000 subjects from 92 datasets. These models address cross-subject and cross-device variability but do not evaluate robustness to preprocessing choices, the variability source we study here.

\paragraph{Domain generalization.}
Methods such as IRM~\citep{arjovsky2019invariant}, GroupDRO~\citep{sagawa2020distributionally}, and DANN~\citep{ganin2016domain} train models invariant to environment-level distribution shifts~\citep{zhou2022domain}, treating environments as opaque, unstructured groups. Unlike these methods, preprocessing interventions provide a \textbf{known compositional graph}: each edge in the Boolean lattice corresponds to toggling exactly one step, enabling edge-level diagnostics and regularization that generic DG methods cannot exploit.


\section{Experimental Framework}
\label{sec:framework}

\subsection{Preprocessing as Structured Intervention}
\label{sec:interventions}

The framework uses $K{=}7$ binary preprocessing interventions, each toggling between two commonly used options selected based on their documented impact on EEG decoding~\citep{kessler2025preprocessing,delpup2025more}:

\begin{table}[ht]
\centering
\caption{Seven atomic preprocessing interventions. Each has two options; all $2^7{=}128$ combinations are evaluated. ``Impact'' indicates prior evidence of effect on decoding performance.}
\label{tab:interventions}
\small
\begin{tabular}{@{}clllc@{}}
\toprule
\# & Intervention & Option A (0) & Option B (1) & Impact \\
\midrule
$a_1$ & Reference & Original & Common avg.\ ref. & Medium \\
$a_2$ & High-pass filter & 0.1\,Hz & 0.5\,Hz & High \\
$a_3$ & Low-pass filter & 45\,Hz & 30\,Hz & High \\
$a_4$ & Baseline correction & None & 200\,ms subtractive & High \\
$a_5$ & Artifact attenuation & Off & ASR~\citep{chang2020eeglab} ($\tau{=}20$) & High \\
$a_6$ & Epoch rejection & Off & Autoreject~\citep{jas2017autoreject} & Med-High \\
$a_7$ & Bad-channel repair & Off & RANSAC + interp. & Medium \\
\bottomrule
\end{tabular}
\end{table}

The 128 pipelines form a Boolean lattice, a 7-dimensional hypercube $\{0,1\}^7$, where each node is a pipeline and each edge connects two pipelines differing by exactly one intervention. This lattice has 448 undirected edges and diameter~7. For each raw EEG recording, we apply all 128 pipelines, producing 128 counterfactual versions of the same neural data.

\subsection{Metrics}

\paragraph{Counterfactual Flip Rate (CFR).}
For a trained model $f_\theta$ and a raw trial $x_i$ with label $y_i$, let $\hat{y}_i^{(p)} = \arg\max f_\theta(p(x_i))$ denote the predicted class under pipeline $p$. The CFR measures how often predictions change:
\begin{equation}
    \text{CFR} = \frac{1}{N \cdot |\mathcal{P}|} \sum_{i=1}^{N} \sum_{p \in \mathcal{P}} \mathbf{1}\!\left[\hat{y}_i^{(p)} \neq \hat{y}_i^{(p_0)}\right]
\end{equation}
where $p_0$ is a reference pipeline and $\mathcal{P}$ is the set of all 128 pipelines. We also report $\text{MaxCFR} = \max_{p,q} \frac{1}{N}\sum_i \mathbf{1}[\hat{y}_i^{(p)} \neq \hat{y}_i^{(q)}]$.

\paragraph{Per-intervention effect.}
For each intervention $a_k$, we compute the average accuracy change when toggling $a_k$ while holding all other interventions fixed:
\begin{equation}
    \Delta_k = \frac{1}{2^{K-1}} \sum_{p:\, p_k=0} \left[\text{Acc}(p \oplus e_k) - \text{Acc}(p)\right]
\end{equation}
where $p \oplus e_k$ denotes the pipeline obtained by flipping bit $k$. The absolute effect $|\Delta_k|$ measures the magnitude of sensitivity to intervention $k$, averaged over the 64 pipeline pairs that differ only on $a_k$.

\subsection{Datasets}

We use six publicly available datasets spanning distinct EEG paradigms:

\begin{table}[ht]
\centering
\caption{Dataset summary. All models use EEGNet-v4~\citep{lawhern2018eegnet} with 3-fold subject-wise cross-validation.}
\label{tab:datasets}
\footnotesize
\begin{tabular}{@{}lcccccl@{}}
\toprule
Dataset & Task & Cls & Subj & Ch & Epoch & Source \\
\midrule
BCI-IV-2a~\citep{brunner2008bci} & MI & 4 & 9 & 22 & 4.0\,s & MOABB~\citep{jayaram2018moabb} \\
PhysionetMI~\citep{schalk2004bci2000} & MI & 2 & 20 & 64 & 3.0\,s & MOABB~\citep{jayaram2018moabb} \\
Sleep-EDF~\citep{kemp2000analysis} & Sleep & 5 & 15 & 2 & 30.0\,s & PhysioNet~\citep{goldberger2000physiobank} \\
BNCI2014-009 & P300 & 2 & 10 & 16 & 0.8\,s & MOABB~\citep{jayaram2018moabb} \\
Lee2019-ERP & P300 & 2 & 10 & 62 & 1.0\,s & MOABB~\citep{jayaram2018moabb} \\
SEED-IV & Emotion & 4 & 15 & 62 & 4.0\,s & BCMI~\citep{zheng2019emotionmeter} \\
\bottomrule
\end{tabular}
\end{table}

\paragraph{Protocol.}
\textbf{Sensitivity analysis} (Section~\ref{sec:results}): train EEGNet on a single anchor pipeline $p_0$, evaluate on all 128 pipelines to isolate preprocessing sensitivity from training effects. Per-intervention effects $\Delta_k$ average over 64 pipeline pairs and are anchor-independent. \textbf{Mitigation} (Section~\ref{sec:napgi}): NA-PGI and all baselines train on all pipeline views with matched budgets (50 epochs, same optimizer/backbone, 3-fold CV split before preprocessing) and are evaluated on all 128 pipelines.
This protocol is not intended to model arbitrary test-time distribution shift. Instead, it isolates an analyst-induced counterfactual: the same raw EEG trial may be processed by different defensible pipelines across laboratories, software defaults, or deployment sites. CFR measures whether a decoder's prediction is stable under these reasonable analytical choices while holding the raw signal, label, and model fixed.

\paragraph{Computational cost.}
All preprocessing uses MNE-Python~\citep{gramfort2013mne}; generating 128 variants takes ${\sim}10$ min/subject (CPU), and ERM training takes ${\sim}1$ GPU-hour/dataset on an A100.


\section{Results}
\label{sec:results}

\subsection{Preprocessing Sensitivity is Real and Task-Specific}

Table~\ref{tab:cfr} summarizes baseline sensitivity across all six datasets. On motor imagery (BCI-IV-2a), a mean of \textbf{42.4\% of predictions flip} across 128 pipelines. Emotion recognition (SEED-IV) shows similarly high sensitivity (35.8\%). Even on high-accuracy tasks, sensitivity is nontrivial: 9.6\% for sleep staging and 2.6--4.1\% for P300/ERP.

\begin{table}[t]
\centering
\caption{Preprocessing sensitivity across tasks (ERM-single, EEGNet, 3-fold CV). CFR = mean fraction of test trials whose prediction changes across 128 pipelines.}
\label{tab:cfr}
\small
\begin{tabular}{@{}lccc@{}}
\toprule
Dataset & Channels & Mean Acc.\ (\%) & CFR (\%) \\
\midrule
BCI-IV-2a (MI, 4-cls) & 22 & 37.6 & \textbf{42.4} \\
SEED-IV (Emotion, 4-cls) & 62 & 31.9 & 35.8 \\
PhysionetMI (MI, 2-cls) & 64 & 57.7 & 21.7 \\
Sleep-EDF (Sleep, 5-cls) & 2 & 85.6 & 9.6 \\
Lee2019-ERP (ERP, 2-cls) & 62 & 84.1 & 4.1 \\
P300 (ERP, 2-cls) & 16 & 83.4 & 2.6 \\
\bottomrule
\end{tabular}
\end{table}

\begin{figure}[t]
    \centering
    \includegraphics[width=0.75\linewidth]{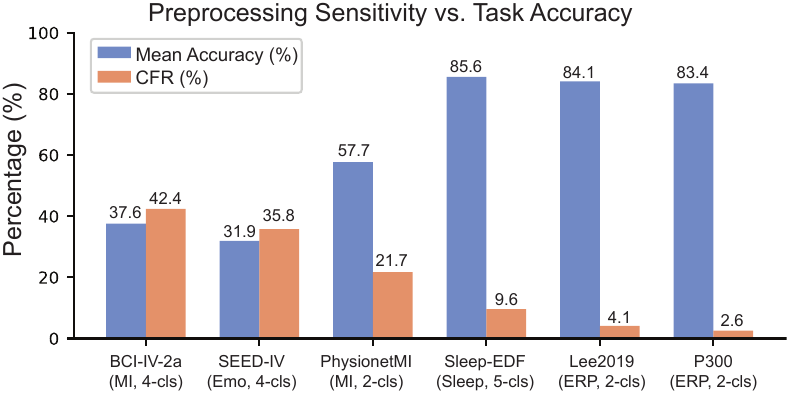}
    \caption{Preprocessing sensitivity (CFR) inversely correlates with task accuracy across all six datasets. Preprocessing choices matter most when the model is least confident.}
    \label{fig:cfr_summary}
\end{figure}

Sensitivity inversely correlates with task accuracy (Figure~\ref{fig:cfr_summary}): when the decoder is least confident (BCI-IV-2a, 37.6\%), preprocessing has the most influence. The best-worst pipeline gap reaches 24.0 percentage points on BCI-IV-2a (49.7\% vs.\ 25.7\%), meaning the choice of preprocessing alone determines nearly a quarter of the accuracy range.

\begin{figure}[t]
    \centering
    \includegraphics[width=0.65\linewidth]{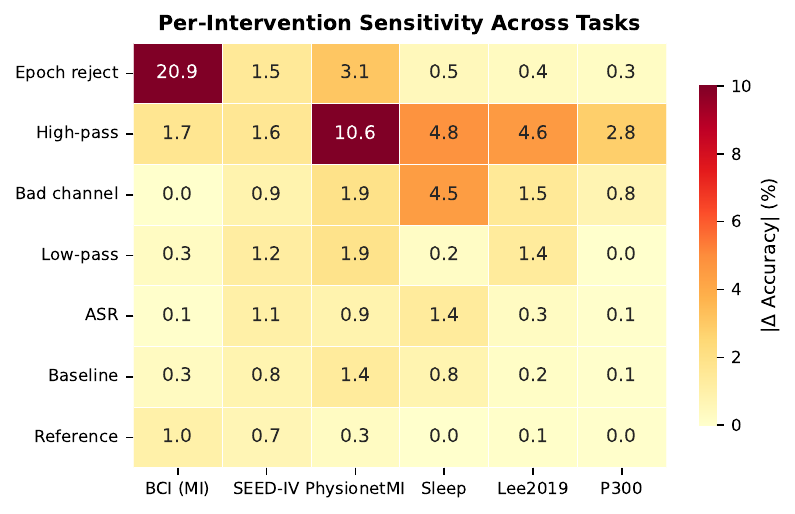}
    \caption{\textbf{Preprocessing sensitivity is task-specific.} Each cell shows the mean absolute per-pair accuracy change $\mathbb{E}[|\delta|]$ (\%) when toggling one intervention across 64 pipeline pairs. Color scale clipped at 10\% for readability; epoch rejection on BCI-IV-2a reaches 20.9\%. Signed effects in Appendix Figure~\ref{fig:signed}.}
    \label{fig:heatmap}
\end{figure}

Which interventions drive this sensitivity differ across tasks (Figure~\ref{fig:heatmap}): epoch rejection dominates BCI-IV-2a ($|\Delta|{=}20.9\%$), while high-pass filtering dominates Sleep-EDF ($4.8\%$) and P300 ($2.8\%$). Pairwise Spearman rank correlations of intervention importance are low (mean $\rho{=}0.274$; Appendix Table~\ref{tab:spearman}), confirming that rankings are largely non-transferable across tasks.

\subsection{Sensitivity is Practically Near-Additive under Binary Interventions}

Prior multiverse studies report marginal effects but do not characterize whether preprocessing steps interact. We apply a Walsh-Hadamard decomposition~\citep{beauchamp1975walsh} to the $2^7$ accuracy hypercube, partitioning total variance into main effects (order 1), pairwise interactions (order 2), and higher-order terms.

\begin{table}[t]
\centering
\caption{Walsh-Hadamard variance decomposition of per-pipeline accuracy across all six datasets. In absolute terms, interactions (order $\geq$2) contribute $\leq$0.2\% of total variance; however, when measured relative to non-mean variance, interaction shares range from 0.6\% (BCI) to 54\% (SEED-IV). Despite this, greedy step-by-step optimization achieves accuracy within 2.5\% of the oracle on all six datasets (see text).}
\label{tab:wht}
\small
\begin{tabular}{@{}lcccc@{}}
\toprule
Dataset & Order 0 (mean) & \textbf{Order 1 (main)} & Order 2 (pairs) & Order 3+ \\
\midrule
BCI-IV-2a & 92.8\% & \textbf{7.2\%} & $<$0.1\% & $<$0.01\% \\
PhysionetMI & 99.0\% & \textbf{0.9\%} & 0.1\% & $<$0.1\% \\
SEED-IV & 99.7\% & \textbf{0.1\%} & 0.1\% & 0.1\% \\
Sleep-EDF & 99.8\% & \textbf{0.2\%} & $<$0.1\% & $<$0.01\% \\
Lee2019-ERP & 99.9\% & \textbf{0.1\%} & $<$0.1\% & $<$0.01\% \\
P300 & 100.0\% & \textbf{$<$0.1\%} & $<$0.01\% & $<$0.01\% \\
\bottomrule
\end{tabular}
\end{table}

The result (Table~\ref{tab:wht}) shows that across all six datasets, \textbf{interactions contribute $\leq$0.2\% of total variance in absolute terms}. However, because the order-0 mean dominates total variance (92--100\%), the picture changes when measured relative to non-mean variance: the interaction share of non-mean variance ranges from 0.6\% on BCI-IV-2a to 54\% on SEED-IV. On SEED-IV, main effects and interactions are comparable in magnitude once the grand mean is removed.

The practical question is whether these interactions are large enough to require joint optimization. We validate this with a greedy step-by-step optimization experiment across all six datasets: the greedy-optimal pipeline achieves accuracy within 2.5\% of the oracle best-of-128 on every dataset, with SEED-IV showing the largest gap (2.5\%) and Lee2019-ERP the smallest (0.4\%). \textbf{Step-wise tuning is thus a strong practical approximation under the binary intervention design studied here}, tuning each intervention in isolation captures most of the achievable improvement, even on datasets where the relative interaction share is substantial. This additivity may not extend to continuous preprocessing parameters (e.g., HPF cutoff sweeps), where~\citet{kessler2025preprocessing} reported meaningful interactions.

Exploratory signal-level analysis suggests candidate mechanistic pathways for why specific interventions matter: high-pass filtering correlates with sleep staging accuracy through kurtosis reduction, while epoch rejection affects motor imagery through training-set composition changes (full analysis in Appendix~\ref{app:signal}).

\subsection{Pipeline Disagreement as Epistemic Uncertainty}

Given that preprocessing changes predictions without changing the data, a natural question is whether this instability can be quantified per trial. Preprocessing Uncertainty (PU) for a trial $x_i$ is defined as:
$\text{PU}(x_i) = 1 - \max_{c} \frac{1}{|\mathcal{P}|}\sum_{p} \mathbf{1}[\hat{y}_i^{(p)} = c]$,
ranging from 0 (all pipelines agree) to $1-1/|\mathcal{C}|$ (uniform disagreement). PU captures analyst degrees of freedom rather than model stochasticity.

We evaluate PU as an error detector using standard UQ benchmarks (for a comprehensive review of uncertainty quantification techniques, see~\citet{abdar2021review}). Table~\ref{tab:pu} summarizes PU error-detection performance and its correlation with model-based uncertainty across three datasets. On PhysionetMI, PU achieves AUROC 0.712 for error detection, comparable to softmax entropy (0.764) and MC Dropout (0.721). PU correlates only moderately with model-based methods (mean $\rho{=}0.40$ vs.\ softmax, $\rho{=}0.33$ vs.\ MC Dropout across three datasets), substantially lower than the mutual correlation between softmax and MC Dropout ($\rho{=}0.55$). The degree of complementarity is dataset-dependent: on BCI-IV-2a, PU is nearly independent ($\rho{=}0.30$), while on PhysionetMI, correlation is moderate ($\rho{=}0.52$). On PhysionetMI, combining PU with softmax yields AUROC 0.783, exceeding either alone and demonstrating complementarity; on BCI-IV-2a, the combination does not improve over softmax alone, suggesting that PU's value is greatest when preprocessing sensitivity is high. Full calibration and selective prediction results are in the appendix.

\begin{table}[t]
\centering
\caption{PU as error detector: AUROC for error detection and Spearman correlation with model-based uncertainty across three datasets (mean over 3 folds).}
\label{tab:pu}
\small
\begin{tabular}{@{}lcccccc@{}}
\toprule
Dataset & PU & Softmax & MC & $\rho$(PU, Soft) & $\rho$(PU, MC) & $\rho$(Soft, MC) \\
\midrule
BCI-IV-2a & 0.548 & 0.636 & 0.604 & 0.298 & 0.283 & 0.625 \\
PhysionetMI & 0.712 & 0.764 & 0.721 & 0.519 & 0.504 & 0.695 \\
SEED-IV & 0.528 & 0.525 & 0.538 & 0.379 & 0.189 & 0.334 \\
\bottomrule
\end{tabular}
\end{table}

Oracle pipeline selection reveals substantial performance headroom: on BCI-IV-2a, keeping only the top 50\% of pipelines improves accuracy by +10.4\% (Appendix Table~\ref{tab:selective}). Combined with the additivity result, this implies that a practitioner who tunes each preprocessing step independently on a validation set should approach the oracle bound.


\section{Mitigation: Normalized Adaptive PGI}
\label{sec:napgi}

Given that preprocessing sensitivity is real, additive, and task-specific, we ask: can we train decoders that are less sensitive to preprocessing choices? We propose Normalized Adaptive Pipeline Generator Invariance (NA-PGI), a plug-in training objective that penalizes prediction drift across atomic preprocessing changes with two key innovations for cross-dataset robustness.

\subsection{Base Method: Pipeline Generator Invariance}

The 128 pipelines form a Boolean lattice where each edge toggles exactly one of the $K{=}7$ atomic interventions. This graph-structured regularization is inspired by the observation that preprocessing interventions have a natural compositional structure: each edge in the lattice corresponds to toggling exactly one preprocessing step, making edge-level consistency a minimal and interpretable invariance constraint. PGI adds a consistency loss on these edges: for centered logits $\bar{z}_i^{(p)} = z_i^{(p)} - \frac{1}{C}\sum_c z_{i,c}^{(p)}$, the base PGI loss is:
\begin{equation}
    \mathcal{L}_{\text{PGI}} = \frac{1}{|E|} \sum_{(p,q) \in E} w_{(p,q)} \cdot \frac{1}{B}\sum_{i=1}^{B}\|\bar{z}_i^{(p)} - \bar{z}_i^{(q)}\|_2^2
\end{equation}

\subsection{Problem: Loss Scale Mismatch}

Na\"ive PGI with a fixed $\lambda$ fails across datasets because \textbf{logit magnitudes vary by up to 50$\times$} depending on channel count and task complexity. A $\lambda$ tuned for BCI-IV-2a (22 channels, small logits) causes collapse on SEED-IV (62 channels, large logits), as the PGI penalty overwhelms the supervised loss.

\subsection{Solution: NA-PGI}

We introduce two complementary fixes, each addressing a different failure mode:

\paragraph{Fix 1: Loss normalization (scale invariance).}
We divide the PGI loss by the detached logit variance, making the penalty scale-invariant:
\begin{equation}
    \mathcal{L}_{\text{N-PGI}} = \frac{\mathcal{L}_{\text{PGI}}}{\text{sg}[\text{Var}(\bar{z})] + \epsilon}
\end{equation}
where $\text{sg}[\cdot]$ denotes stop-gradient (preventing the model from inflating logits to trivially minimize the ratio). This ensures that $\lambda{=}1$ carries the same semantic meaning regardless of dataset: ``the invariance penalty should be comparable in magnitude to the cross-entropy loss.''

\paragraph{Fix 2: Adaptive $\lambda$ (collapse prevention).}
We modulate $\lambda$ based on the running evaluation CFR:
\begin{equation}
    \lambda_{\text{eff}} = \lambda \cdot \text{clamp}\!\left(\frac{\overline{\text{CFR}}}{\tau}, 0.01, 5.0\right)
\end{equation}
where $\overline{\text{CFR}}$ is an exponential moving average of the validation CFR and $\tau{=}0.15$ is a target CFR. When CFR is high (model is sensitive), $\lambda$ is boosted; when CFR drops toward zero (collapse risk), $\lambda$ is automatically reduced, allowing the supervised loss to recover. The EMA is initialized to $\tau$ so that $\lambda$ starts at its base value.

\paragraph{Why both fixes are necessary.} Our ablation (Table~\ref{tab:ablation}) shows that normalization alone still collapses on 2/3 datasets (the gradient landscape remains unstable without the dynamic safety net), while adaptive-only without normalization yields inconsistent results across datasets (the raw loss scale still varies). Only the combination provides robust, cross-dataset performance with a single $\lambda{=}1$.

\subsection{Results}

We compare NA-PGI against six baselines on three high-CFR datasets (Table~\ref{tab:main}).

\begin{table}[t]
\centering
\caption{CFR (fraction) across methods on 128-pipeline evaluation. Domain generalization baselines in our simplified implementations (GroupDRO, IRM) provide marginal improvement, consistent with DomainBed~\citep{gulrajani2021search}. NA-PGI achieves the best average CFR reduction.}
\label{tab:main}
\small
\begin{tabular}{@{}lccc|c@{}}
\toprule
Method & BCI-IV-2a & PhysionetMI & SEED-IV & Avg.\ $\Delta$ \\
\midrule
ERM-single & 0.424 & 0.217 & 0.358 & --- \\
ERM-mixed & 0.424 ($-$0\%) & 0.221 ($-$2\%) & 0.323 (+10\%) & +3\% \\
GroupDRO & 0.412 (+3\%) & 0.223 ($-$3\%) & 0.332 (+7\%) & +2\% \\
IRM & 0.412 (+3\%) & 0.228 ($-$5\%) & 0.314 (+12\%) & +3\% \\
Consistency & 0.418 (+2\%) & \textbf{0.178} (+18\%) & 0.311 (+13\%) & +11\% \\
CORAL & 0.419 (+1\%) & 0.252 ($-$16\%) & 0.274 (+24\%) & +3\% \\
\textbf{NA-PGI} & \textbf{0.406} (+4\%) & 0.187 (+14\%) & \textbf{0.233} (+35\%) & \textbf{+18\%} \\
\bottomrule
\end{tabular}
\end{table}

Our analysis yields three insights regarding domain generalization under preprocessing shifts:

\paragraph{The DomainBed phenomenon in EEG.}
Consistent with large-scale DG meta-analyses~\citep{gulrajani2021search}, our simplified implementations of GroupDRO and IRM yield only +2--3\% average improvement over ERM-mixed. Canonical implementations with stable group tracking may perform differently (see Appendix for implementation details).

\paragraph{Feature-space alignment instability.}
Our simplified CORAL~\citep{sun2016deep} implementation exhibits instability: +24\% on SEED-IV but $-$16\% on PhysionetMI. We hypothesize that enforcing feature-space alignment may be overly aggressive for EEG, though this may also reflect our simplified implementation (feature-mean divergence rather than full covariance).

\paragraph{Prediction-space consistency vs.\ NA-PGI.}
In contrast, regularizing the output prediction space proves much safer. Consistency regularization, which treats all 128 pipelines as independent, equally distant domains, achieves a robust +11\% average, particularly on PhysionetMI (+18\%). NA-PGI achieves the highest overall efficacy (+18\% average), largely driven by a substantial +35\% on SEED-IV. While Consistency treats pipelines as a flat set, NA-PGI leverages the topological structure of the intervention lattice, penalizing drift along atomic edges rather than across arbitrary pipeline pairs. Combined with adaptive scale normalization, this achieves strong invariance without sacrificing stability on high-dimensional data. NA-PGI uses a single $\lambda{=}1$ without per-dataset tuning.

\paragraph{Multi-seed stability (5 seeds).} On PhysionetMI (64 channels), NA-PGI is highly stable across 5 random seeds (CFR $0.135 \pm 0.037$), with all seeds showing substantial improvement over ERM (0.217). SEED-IV (62 channels) achieves the largest mean reduction ($0.114 \pm 0.065$ vs.\ ERM 0.358, $-68\%$), though variability is high and one seed exhibits fold-level collapse. On BCI-IV-2a (22 channels, 4-class), improvement is marginal ($0.410 \pm 0.056$ vs.\ ERM 0.424, $-3\%$) with one seed showing fold-level collapse. We analyze the failure mode in Section~\ref{sec:discussion}.

\paragraph{Accuracy-CFR tradeoff.} Table~\ref{tab:acc_cfr} reports both accuracy and CFR for the main methods. NA-PGI maintains accuracy within 1 percentage point of ERM on PhysionetMI and SEED-IV while reducing CFR by 14--35\%. On BCI-IV-2a, the accuracy cost is larger ($-$7.1\,pp), reflecting the representational capacity constraint on low-channel data. This tradeoff may be acceptable in applications where preprocessing stability is prioritized, but it marks low-channel EEG as a challenging regime for invariance training.

\begin{table}[t]
\centering
\caption{Mean accuracy (\%) and CFR (\%) for key methods. NA-PGI achieves the strongest CFR reduction on high-density datasets, while exposing a substantial accuracy-robustness tradeoff on low-channel BCI-IV-2a.}
\label{tab:acc_cfr}
\small
\begin{tabular}{@{}l|cc|cc|cc@{}}
\toprule
 & \multicolumn{2}{c|}{BCI-IV-2a} & \multicolumn{2}{c|}{PhysionetMI} & \multicolumn{2}{c}{SEED-IV} \\
Method & Acc & CFR & Acc & CFR & Acc & CFR \\
\midrule
ERM-single & 37.6 & 42.4 & 57.7 & 21.7 & 31.9 & 35.8 \\
Consistency & 38.8 & 41.8 & 57.1 & 17.8 & 32.3 & 31.1 \\
NA-PGI & 30.5 & 40.6 & 56.9 & 18.7 & 33.7 & 23.3 \\
\bottomrule
\end{tabular}
\end{table}

\begin{table}[t]
\centering
\caption{Ablation study. Both normalization and adaptive $\lambda$ are necessary. Normalize-only collapses on 2/3 datasets; adaptive-only is inconsistent across datasets. Only NA-PGI (both) provides robust cross-dataset improvement.}
\label{tab:ablation}
\small
\begin{tabular}{@{}cclccc@{}}
\toprule
Norm. & Adapt. & Config & BCI & Phys. & SEED \\
\midrule
\ding{55} & \ding{55} & Old PGI ($\lambda{=}5$) & 0.337 & 0.235 & collapse \\
\ding{51} & \ding{55} & Normalize-only & 0.406 & 0.214 & collapse \\
\ding{55} & \ding{51} & Adaptive-only & 0.395 & 0.126 & 0.306 \\
\ding{51} & \ding{51} & \textbf{NA-PGI} & 0.406 & 0.187 & \textbf{0.233} \\
\bottomrule
\end{tabular}
\end{table}


\section{Discussion}
\label{sec:discussion}

\paragraph{Relation to concurrent work.}
\citet{kessler2025preprocessing} reported meaningful interactions for continuous preprocessing parameters in ERP decoding; our binary design finds interactions small in absolute terms ($\leq$0.2\%) but up to 54\% of non-mean variance on SEED-IV, the practical implication is identical: validate preprocessing choices for each task. \citet{delpup2025more} compared preprocessing levels across six tasks; we decompose individual steps and explain why they matter. \citet{delorme2023eeg} showed that ERP significance benefits little from complex preprocessing; we extend this to deep learning decoders and provide both a diagnostic (PU) and a mitigation method (NA-PGI). The underspecification framework of \citet{damour2022underspecification} provides a complementary perspective: our 128 pipelines represent a concrete instance of the ``pipeline multiplicity'' problem, where many equally valid preprocessing choices lead to divergent predictions. Our simplified implementations of GroupDRO, IRM, and CORAL (Table~\ref{tab:main}) provide at most 3\% average CFR reduction, echoing DomainBed's~\citep{gulrajani2021search} observation that ERM is hard to beat. One structural advantage of NA-PGI is that it exploits the compositional structure of preprocessing interventions through edge-level consistency, rather than treating pipelines as opaque domains; recent DG surveys~\citep{zhou2022domain} suggest that domain structure, when available, should be incorporated into invariance constraints, and our Boolean lattice provides exactly this structure.

\paragraph{Scope conditions.}
Three boundaries define the regime where our findings and methods apply.
\textbf{Channel density as a scope condition for invariance training.} These results identify channel density as an important scope condition for preprocessing-invariant training. In high-density EEG ($\geq$60 channels), redundancy across channels allows the model to preserve task-relevant information while enforcing pipeline consistency. In low-channel EEG (e.g., 22-channel BCI-IV-2a), invariance constraints can over-compress sparse discriminative signals, making softer consistency objectives (e.g., Consistency regularization) preferable. This tradeoff is not a limitation of NA-PGI per se, but a structural property of the invariance-capacity balance.
\textbf{Binary vs.\ continuous intervention design.} The Walsh-Hadamard additivity analysis operates on binary intervention choices (e.g., ``apply ASR or not''). Continuous preprocessing parameters such as precise bandpass cutoffs or artifact rejection thresholds may introduce non-linear effects that our binary factorial design does not capture, as suggested by~\citet{kessler2025preprocessing}. Extending to continuous parameter sweeps is a natural next step.
\textbf{Architecture coverage.} All results use EEGNet and ShallowNet. Sensitivity patterns may differ for larger models or EEG foundation models~\citep{kostas2021bendr,jiang2024large,csbrain2025,neuript2025,reve2025,wang2026neurostorm}; applying our CFR framework to these models would test whether large-scale pretraining reduces or eliminates preprocessing sensitivity.

\paragraph{Cross-architecture validation.}
To verify that preprocessing sensitivity is not an artifact of EEGNet, we repeat the ERM-single evaluation with ShallowNet~\citep{schirrmeister2017deep} (Table~\ref{tab:shallow}). CFR values are consistent across architectures (within 6 percentage points), confirming that CFR reflects data-level variability rather than architecture-specific artifacts.

\begin{table}[ht]
\centering
\caption{CFR and accuracy under ERM-single for EEGNet vs.\ ShallowNet on three high-CFR datasets. Preprocessing sensitivity is consistent across architectures.}
\label{tab:shallow}
\small
\begin{tabular}{@{}lcccc@{}}
\toprule
Dataset & \multicolumn{2}{c}{EEGNet} & \multicolumn{2}{c}{ShallowNet} \\
\cmidrule(lr){2-3} \cmidrule(lr){4-5}
 & Acc & CFR & Acc & CFR \\
\midrule
BCI-IV-2a & 37.6 & 42.4 & 33.0 & 47.8 \\
PhysionetMI & 57.7 & 21.7 & 56.8 & 22.7 \\
SEED-IV & 31.9 & 35.8 & 31.5 & 37.5 \\
\bottomrule
\end{tabular}
\end{table}

\paragraph{Practical recommendations.}
\begin{enumerate}[nosep,leftmargin=*]
\item \textbf{Optimize preprocessing step-by-step.} Greedy step-by-step tuning achieves accuracy within 2.5\% of the oracle on all six datasets; exhaustive pipeline search is unnecessary.
\item \textbf{Consider NA-PGI for high-channel recordings.} NA-PGI provides out-of-the-box preprocessing robustness with $\lambda{=}1$ on 60+ channel setups. For low-channel configurations, Consistency regularization is a safer alternative.
\item \textbf{Report PU alongside accuracy.} A model with 80\% accuracy and 40\% CFR is fundamentally different from one with 80\% accuracy and 2\% CFR; PU makes this distinction visible at zero additional training cost.
\end{enumerate}

\paragraph{Future work.}
Natural extensions include continuous parameter sweeps (e.g., HPF cutoff), applying CFR as a robustness diagnostic for EEG foundation models~\citep{csbrain2025,neuript2025,reve2025}, and combining PU with model-based uncertainty in a principled Bayesian framework.

\paragraph{Broader impact.}
Preprocessing sensitivity has immediate implications for clinical EEG, where brain-computer interfaces~\citep{blankertz2006bci} and seizure detection rely on consistent predictions.

\section{Conclusion}

By formalizing preprocessing choices as a counterfactual intervention space, we showed that EEG predictions are surprisingly unstable: up to 42\% of trial-level predictions flip across 128 pipelines, a variability invisible to standard uncertainty methods. The Walsh-Hadamard decomposition makes this instability decomposable, PU makes it measurable per trial, and NA-PGI demonstrates that structured regularization can reduce it under clear scope conditions. Rather than prescribing a universally optimal preprocessing pipeline, our results establish that preprocessing sensitivity should be measured, reported, and optimized as a first-class reliability property of EEG decoders.


\bibliographystyle{unsrtnat}
\bibliography{references}

\newpage
\appendix
\section{Additional Results}
\label{app:additional}

\subsection{Spearman Rank Correlations of Intervention Importance}

\begin{table}[ht]
\centering
\caption{Spearman rank correlations of intervention importance on three representative datasets (BCI-IV-2a, Sleep-EDF, P300).}
\label{tab:spearman}
\small
\begin{tabular}{@{}lccc@{}}
\toprule
 & BCI-IV-2a & Sleep-EDF & P300 \\
\midrule
BCI-IV-2a & 1.000 & $-$0.214 & +0.214 \\
Sleep-EDF & & 1.000 & \textbf{+0.821} \\
P300 & & & 1.000 \\
\midrule
\multicolumn{4}{l}{Mean pairwise $\rho = 0.274$} \\
\bottomrule
\end{tabular}
\end{table}

\subsection{Signal-Level Correlates}
\label{app:signal}

For one representative subject per dataset, we compute trial-level signal statistics, amplitude, channel variance dispersion, low-frequency power, trial-level maximum, and kurtosis, over 100 sampled trials and examine how interventions alter these features and how those alterations relate to accuracy.

\begin{table}[ht]
\centering
\caption{Signal-accuracy correlations ($r$) and per-intervention associations on three representative datasets (one representative subject, 100 trials per dataset). Note: these are correlational, not causal.}
\label{tab:mediation}
\small
\begin{tabular}{@{}llcll@{}}
\toprule
Dataset & Top correlate & $r$ & Dominant intervention & Mediating mechanism \\
\midrule
BCI-IV-2a & Amplitude & +0.19 & Epoch rejection & $\Delta$\,lf\_power \\
Sleep-EDF & Ch.\ var.\ disp. & +0.60 & High-pass filter & $\Delta$\,kurtosis ($-$0.46) \\
P300 & Kurtosis & +0.38 & High-pass filter & $\Delta$\,kurtosis ($-$0.12) \\
\bottomrule
\end{tabular}
\end{table}

\begin{itemize}
    \item \textbf{Sleep staging:} Accuracy correlates strongly with channel variance dispersion ($r{=}+0.60$) and kurtosis ($r{=}+0.58$). High-pass filtering at 0.5\,Hz (vs.\ 0.1\,Hz) significantly reduces kurtosis ($\Delta{=}-0.46$), which in turn reduces accuracy by 4.8\%.
    \item \textbf{P300:} Kurtosis correlates with accuracy ($r{=}+0.38$), and high-pass filtering reduces kurtosis ($\Delta{=}-0.12$), reducing accuracy by 2.8\%.
    \item \textbf{Motor imagery:} Epoch rejection operates through a different pathway: removing outlier trials changes the training distribution's low-frequency power. The signal-accuracy correlation is weaker ($r{=}+0.19$), suggesting that for MI, the effect is more about which data is kept than how the signal is transformed.
\end{itemize}

\subsection{Oracle Pipeline Selection}

\begin{table}[ht]
\centering
\caption{Selective pipeline prediction (oracle upper bound) on three representative datasets: accuracy when retaining only the top-$K$\% pipelines by per-pipeline test accuracy.}
\label{tab:selective}
\small
\begin{tabular}{@{}lccccc@{}}
\toprule
Dataset & All 128 & Top 75\% & Top 50\% & Top 25\% & Top 10\% \\
\midrule
BCI-IV-2a & 37.6 & 41.2 & \textbf{48.0} & 49.4 & 49.7 \\
Sleep-EDF & 85.6 & 87.3 & \textbf{88.4} & 89.4 & 89.6 \\
P300 & 83.4 & 84.0 & \textbf{84.8} & 85.0 & 85.3 \\
\bottomrule
\end{tabular}
\end{table}

\subsection{Signed Intervention Effects}

\begin{figure}[ht]
    \centering
    \includegraphics[width=0.65\linewidth]{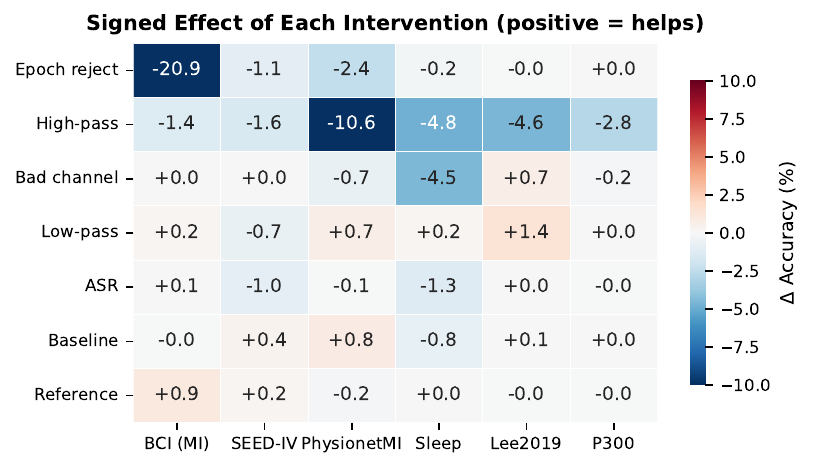}
    \caption{Signed effect ($\Delta_k$, \%) of each intervention across all six datasets. Positive values indicate that enabling the intervention improves accuracy. Note the sign flips for ASR and bad-channel repair across tasks.}
    \label{fig:signed}
\end{figure}

Figure~\ref{fig:signed} shows the signed per-intervention effects, revealing that the same intervention can have opposite effects across tasks. For example, ASR has a weakly positive effect on BCI-IV-2a but a negative effect on Sleep-EDF.

\subsection{PGI Implementation Details}

PGI adds approximately 10 lines to a standard training loop:
{\small
\begin{verbatim}
logits = decoder(views.flatten(0,1)).view(B, V, C)
sup = F.cross_entropy(logits.reshape(B*V,C),
                      y.repeat_interleave(V))
z = logits - logits.mean(dim=-1, keepdim=True)
edge_loss = ((z[:,src]-z[:,dst]).pow(2).sum(-1)*w).mean()
loss = sup + lam * edge_loss
\end{verbatim}
}
where \texttt{src}, \texttt{dst} are pre-computed Hasse edge indices and \texttt{w} contains per-intervention weights. For NA-PGI, add logit-variance normalization (\texttt{/ z.detach().pow(2).mean()}) and adaptive $\lambda$ modulation.

\subsection{NA-PGI Stability Across Seeds}

\begin{table}[ht]
\centering
\caption{NA-PGI CFR across 5 random seeds. PhysionetMI is highly stable (std$=$0.037); SEED-IV shows high variability but consistent improvement over ERM (0.358). BCI-IV-2a shows marginal improvement with occasional fold-level collapse. $^\dagger$At least one fold collapsed (CFR$\to$0); early-stopping may or may not preserve a non-degenerate checkpoint.}
\label{tab:seeds}
\small
\begin{tabular}{@{}lcccccc@{}}
\toprule
Dataset & S42 & S43 & S44 & S45 & S46 & Mean$\pm$Std \\
\midrule
BCI-IV-2a (22ch) & 0.406 & 0.460 & 0.438 & 0.442 & 0.304$^\dagger$ & 0.410$\pm$0.056 \\
PhysionetMI (64ch) & 0.187 & 0.140 & 0.155 & 0.079 & 0.114 & 0.135$\pm$0.037 \\
SEED-IV (62ch) & 0.233 & 0.129 & 0.071 & 0.058$^\dagger$ & 0.077 & 0.114$\pm$0.065 \\
\bottomrule
\end{tabular}
\end{table}

\subsection{NA-PGI Training Dynamics}

\begin{figure}[ht]
    \centering
    \includegraphics[width=0.85\linewidth]{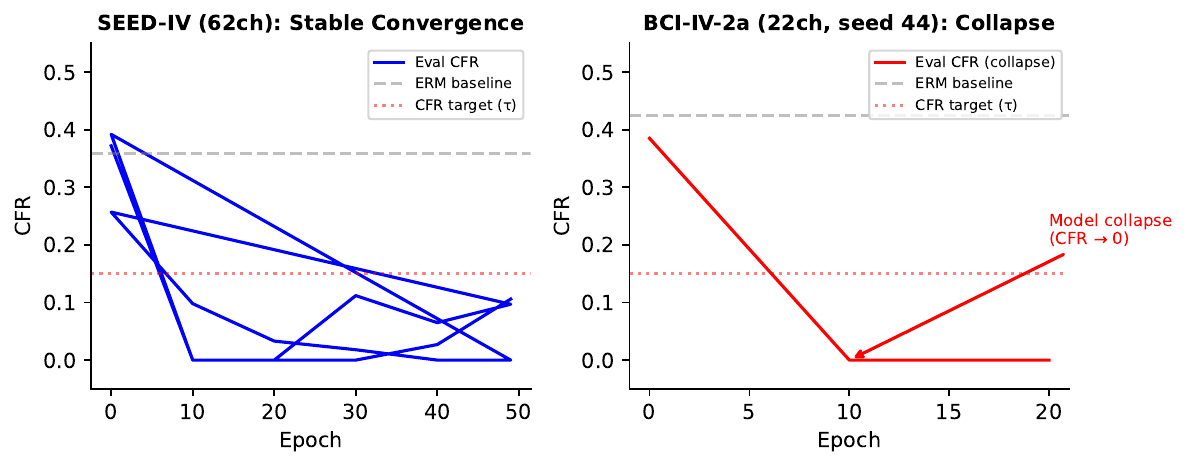}
    \caption{NA-PGI training dynamics. \textbf{Left:} SEED-IV (62ch) shows stable CFR reduction. \textbf{Right:} BCI-IV-2a (22ch, seed 44) exhibits mid-training collapse despite adaptive $\lambda$.}
    \label{fig:dynamics}
\end{figure}

Figure~\ref{fig:dynamics} illustrates the training dynamics of NA-PGI on two contrasting datasets. On SEED-IV (62 channels), the adaptive $\lambda$ mechanism produces stable convergence: CFR decreases smoothly and the model maintains discriminative accuracy. On BCI-IV-2a (22 channels, seed 44), CFR drops abruptly to near zero mid-training, indicating representational collapse. Early-stopping preserves a non-degenerate checkpoint, but the resulting model is worse than the ERM baseline.

\subsection{Training Budget Comparability}

All multi-view methods (ERM-mixed, Consistency, GroupDRO, IRM, CORAL, NA-PGI) use the same number of pipeline views per batch (8 views for baselines, 256 sampled edges for NA-PGI corresponding to ${\sim}40$ unique views). Total training epochs (50) and optimizer (AdamW, lr=$10^{-3}$) are identical across methods. NA-PGI requires ${\sim}3\times$ wall-clock time due to the larger effective batch from edge sampling.

\subsection{DG Baseline Implementation Notes}

Our GroupDRO implementation uses a reweighting vector over pipeline views within each batch, but because views are randomly subsampled per batch, stable group identities are not maintained across iterations. Our CORAL implementation penalizes feature-mean divergence across pipeline views rather than full covariance alignment. These are simplified variants; canonical implementations with stable group tracking and full covariance penalties may yield different results.


\end{document}